\documentclass[11pt,a4paper]{article}
\usepackage{acl2015}
\usepackage{times}
\usepackage{url}
\usepackage{latexsym}
\usepackage{multirow}
\usepackage{amssymb,amsmath,graphicx,amsfonts,epsfig,epstopdf,datetime}
\usepackage{float}
\usepackage{color}
\usepackage{graphicx}
\usepackage{subfigure}
\title{Learning from LDA using Deep Neural Networks}

\author{Dongxu Zhang$^{1,3}$, Tianyi Luo$^{1,4}$, Dong Wang$^{*1,2}$, Rong Liu$^{1,4}$ \\
  $^1$CSLT, RIIT, Tsinghua University   \\
  $^2$Tsinghua National Lab for Information Science and Technology\\
  $^3$PRIS, Beijing University of Posts and Telecommunications \\
  $^4$Huilan Limited,  Beijing, P.R. China \\
  {\tt {wangdong}99@mails.tsinghua.edu.cn} \\}

\date{}

\begin{document}
\maketitle
\begin{abstract}

  Latent Dirichlet Allocation (LDA) is a three-level hierarchical Bayesian model for topic inference. In spite of its great success, inferring the latent topic distribution with LDA is time-consuming. Motivated by the transfer learning approach proposed by~\newcite{hinton2015distilling},
  we present a novel method that uses LDA to supervise the training of a deep neural network (DNN), so that the DNN can approximate the costly LDA inference with less computation.
  Our experiments on a document classification task show that a simple DNN can learn the LDA behavior pretty well, while the inference is speeded up
  tens or hundreds of times.

\end{abstract}

\section{Introduction}

Probabilistic topic models, for instance Latent Dirichlet Allocation (LDA)~\cite{blei2003latent}, have been extensively studied and
widely used in applications such as topic discovery, document classification and information retrieval. Most of the successful
probabilistic topic models are based on Bayesian networks~\cite{hofmann1999probabilistic,teh2006hierarchical}, where
the random variables and the dependence among them are carefully designed by people and so hold clear meanings in physics and/or statistics.
For this reason, Bayesian topic models can represent the document generation process well and have attained much success
in semantic analysis and related research.

A particular problem of Bayesian topic models, however, is that when the model structure
is complex, the inference for the latent topic distribution (topic mixture weights) is
often untractable. Various approximation methods
have been proposed, such as the variational approach and the sampling method, though the inference is still very slow.

%Though promising, the form of posterior distribution when inferring the hidden variables is complex. So one has to resort to slow or inaccurate approximations methods. %{\color{red}{to compute the posterior distribution over topics}}. Thus the expensive computation cost is an unavoidable problem to be solved.

%can be viewed as conditionally independent hierarchical models in which latent topic variables   previous layer and directly connect to observed variables, words in a document for instance.

Recently, \newcite{hinton2015distilling} proposed a transfer learning approach. In this approach,
a complex model is used as a teacher model to supervise the training of a simpler model. The original proposal
used a complex deep neural network (DNN) to train a simple shallow neural network and obtained performance very
close to the complex DNN.
This motivated our current research that attempts to use a Bayesian model to supervise the training of a neural model.
Specifically, we use
an LDA as the teacher model to guide the training of a DNN, so that the DNN can approximate the behavior and performance
of LDA. A big advantage of this transfer learning from LDA to DNN is that inference with DNN
is much faster than with LDA. This solves a major difficulty of LDA on large-scale online tasks.

We tested the proposed method on a document classification task. The results show that a simple DNN model
can approximate LDA pretty well and the inference speeds up tens or hundreds of times. Interestingly, a preliminary analysis
shows that by the transfer learning, the DNN model seems can discover topics similar to those learned by LDA,
although this information is not explicitly presented in the transfer learning.

%lower layer to higher layer, which indicates DNN have conduct documents' topics clustering from low layer to higher layer. %Further more, we can use fine-tuning method to get better performances on specific task with neural network structure, which is rather difficult for LDA to do so because of its unsupervised property.%

\section{Related work}

%Probabilistic topic models has gain much attention since 1999. The pLSI model proposed by ~\newcite{hofmann1999probabilistic} posits that a document label $d$ and a word $w_n$ are conditionally independent given an unobserved topic $z$.
%This model attempts to relax the simplifying assumption made in the mixture of unigrams model that each document is generated from only one topic
%However, pLSI is not a well-defined generative model of documents; there is no natural way to use it to assign probability to a previously unseen document.
%Also, The linear growth in parameters suggests that the model is prone to overfitting.

This work develops a neural model to approximate the function of LDA~\cite{blei2003latent}, with a direct goal of a fast inference.
%LDA is a generative probabilistic model and belongs to a large family of Bayesian topic models. LDA assumes that a document involves a mixture of latent topics, and each topic is characterized by a distribution over words.
Compared to the early probabilistic models such as pLSI~\cite{hofmann1999probabilistic}, LDA treats the topic mixture as a latent variable rather than a deterministic parameter. This leads to a full generative model that can deal with new documents, but
also causes much more computation in model inference. The DNN-based LDA approximation presented in this paper attempts to solve this problem.

Our work is also closely related to the deep learning research that was largely initiated by~\newcite{hinton2006fast}. DNN is a popular deep
learning model and is capable of learning complex functions and inferring layer-wise patterns. This work leverages these advantages
and uses DNNs to approximate LDA.
%by learning its mapping function from the primary input (i.e., term frequency)
%to the high-level output (i.e., topic mixture). Interestingly, we find that topics can be discovered
%by DNN automatically with this transfer learning, which in turn demonstrates the power of DNN models.
Note that deep learning has been employed in topic modeling, e.g., the approach based on deep Boltzmann machines (DBM)~\cite{hinton2009replicated,srivastava2013modeling}. The difference of our work is that we focus on approximating
a well-trained Bayesian model using a deep neural model, instead of learning the deep model from scratch.

%And deep learning get remarkable improvement about the performance in many different fields e.g. speech recognition, natural language processing, computer vision and information retrieval. But one of the biggest challenge of deep neural network is that it is hard to figure out the meaning of different neurons.
%Later, Teh et al.~\shortcite{teh2006hierarchical} proposed hierarchical Dirichlet processes(HDP). They considered clustering problems involving multiple groups of data,
%where HDP mixture models are able
%to automatically determine the appropriate number of mixture components needed and
%exhibit sharing of statistical strength across groups by having components shared across
%groups. HDP can infer the number of topics automatically , while LDA requires some method of model selection.

Finally, this research is directly motivated by the dark knowledge distiller model~\cite{hinton2015distilling} that employs the knowledge learned by a complex DNN to guide the training of a simpler DNN, or vice versa~\cite{wang2015recurrent}. In this work, we extend this method to learn a neural model with the supervision of a Bayesian model, which is more ambitious and challenging.

\section{Methods}
\label{sec:model}
For a particular document $d$, LDA takes the term frequency (TF) as the input, denoted by $v(d)$.
The inference task is then to derive the topic mixture $\theta(d)$, which is actually the posterior
probability distribution
that the document belongs to the topics. In tasks such as document clustering or classification,
$\theta(d)$ is a good representation for document $d$, with a low dimensionality and a clear semantic interpretation.

Exact inference with LDA is untractable and so various approximation methods are usually used.
This work chooses the variational inference method
proposed by~\newcite{blei2003latent}, which involves
iterative update of the document and word topic mixtures and hence time-consuming.
The basic idea of the LDA to DNN knowledge transfer learning is to train a DNN model which
can simulate the behavior of LDA inference, but with much less computation. More precisely, the DNN
model learns a mapping function $f(v(d);w)$ such that $f(v(d);w)$ approaches to $\theta(d)$, where $w$ denotes the
parameters of the DNN. Note that $\theta(d)$ is a probability distribution. To approximate such
normalized variables, a softmax function is applied to the DNN output and the cross entropy is used
as the training criterion, given by:

\begin{equation}
\label{eq:jtgr}
\mathcal{L}(w) = - \sum_d \sum_{i = 1}^{K} \theta(d)_i \log  f(v(d);w)_i
\end{equation}

\noindent where $K$ denotes the number of topics and the subscript $i$ indexes the dimension. Once the DNN is
trained, the mapping function $f(v(d);w)$ learns the behavior of the LDA model and can be used to predict $\theta(d)$
for new documents.
Compared to the LDA inference, $f(v(d);w)$ can be computed very fast and hence amiable to large-scale online tasks.

We experimented with two DNN structures: a 2-layer DNN (DNN-2L) that involves one hidden layer, and a 3-layer DNN (DNN-3L)
that involves two hidden layers.
In DNN-2L, the number of hidden units is twice of the output units; in DNN-3L, the number of hidden units are three and two
times of the output units for the first and second hidden layer, respectively.
%For each hidden layer, the number of units is 2 times ,
The hyperbolic function is used as the activation function. The training employs the stochastic
gradient descent (SGD) method, and is implemented based on  Theano~\cite{Bastien-Theano-2012}\footnote{\url{http://deeplearning.net/software/theano/}}.

Note that we have assumed that the topics have been learned already. In fact, learning topics is even slower than
inferring the topic mixtures. For example, the empirical Bayesian method proposed by~\newcite{blei2003latent} involves
an alternative variational EM procedure, which is rather slow. However, since the model training can be conducted
off-line, it is not a big concern for online tasks.

\section{Experiments}

\subsection{Database and experimental setup}

The proposed methods are tested on the document classification task with two datasets. The first dataset is Reuters-21578 and we
follow the `LEWISSPLIT' configure to define the training and test data. The documents are labelled in $55$ classes.\footnote{\url{https://kdd.ics.uci.edu/databases/reuters21578/reuters21578.html}}

The second dataset is 20 Newsgroups collected by Ken Lang, which contains about 20,000 articles evenly distributed over $20$
UseNet discussion groups. These groups correspond to the classes in document classification.\footnote{\url{http://www.cs.cmu.edu/afs/cs.cmu.edu/project/theo-20/www/data/news20.html}}

It has been known that LDA performs better with long documents~\cite{tang2014understanding}. To establish a strong LDA baseline, only long documents
are selected for training and test in this study. Considering that 20 Newsgroups is much larger than Reuters-21578, different selection criteria are used
to choose documents for the two datasets, as shown in Table~\ref{tab:stat}. The table also shows the lexicon
size in the LDA and DNN modeling, which corresponds to the dimensionality of the TF feature. Note that this seemingly tricky data selection
is just for building a strong LDA model for the DNN to learn, rather than intensively selecting a working scenario for the proposed method.
In fact, the DNN learning works well with any LDA teacher model, and the performance of the resultant DNN
largely depends on the quality of the teacher LDA.

\begin{table}[!htb]
\begin{center}
\begin{tabular}{|l|c|c|}
\hline
                                     & Reuters  & 20 News\\
\hline
Document length threshold            &  100     & 300 \\
Training documents                   &  3622    & 6312  \\
Test documents                       &  1705    & 1542 \\
\hline
Word frequency threshold             &  30      & 200  \\
Lexicon size (words)                 &  2388    & 1910 \\
\hline
\end{tabular}
\end{center}
\caption{\label{tab:stat} Data profile of the experimental datasets.}
\end{table}

\subsection{Results}
\label{sec:result}

To evaluate the proposed transfer learning, we compare the classification performance with the document
vectors inferred from the LDA-supervised DNN and the original LDA. The support vector machine (SVM) with a linear kernel
is used as the classifier.
%and the document vectors of the training data shown in Table~\ref{tab:stat} are used to train the SVM.
Since LDA is the teacher model, its performance can be regarded as a upper bound of the DNN learning. Additionally, we choose the
popular principle component analysis (PCA)~\cite{jolliffe2002principal} as another baseline and regard it
as a low bound of the learning. All these three methods generate low-dimensional document vectors and are comparable in the sense
of dimension reduction. Note that in many cases LDA does not outperform PCA, though it is not the focus of our study. What we
are concerned with is that in the case where LDA is superior to PCA, the learned DNN can keep this superiority, but with much less computation
cost.

\subsubsection{Document classification}

%\begin{figure}[!htb]
%\centering
%\epsfig{figure=reuters_trend.eps,width=7cm,height=5cm}
%\caption{The classification accuracy of PCA, LDA, DNN (two layers) and DNN (three layers) on the Reuters-21578 dataset.
%The number of the topics varies from 10 to 70. }
%\label{fig:reuters_trend}
%\end{figure}

%\begin{figure}[!htb]
%\centering
%\epsfig{figure=20news_trend.eps,width=7cm,height=5cm}
%\caption{The classification accuracy of PCA, LDA, DNN (two layers) and DNN (three layers) on the 20 Newsgroups dataset.
%The number of the topics varies from 10 to 70.}
%\label{fig:20news_trend}
%\end{figure}

\begin{figure}[!htb]
\centering
\subfigure{
\begin{minipage}[b]{0.4\textwidth}
\includegraphics[width=1\textwidth]{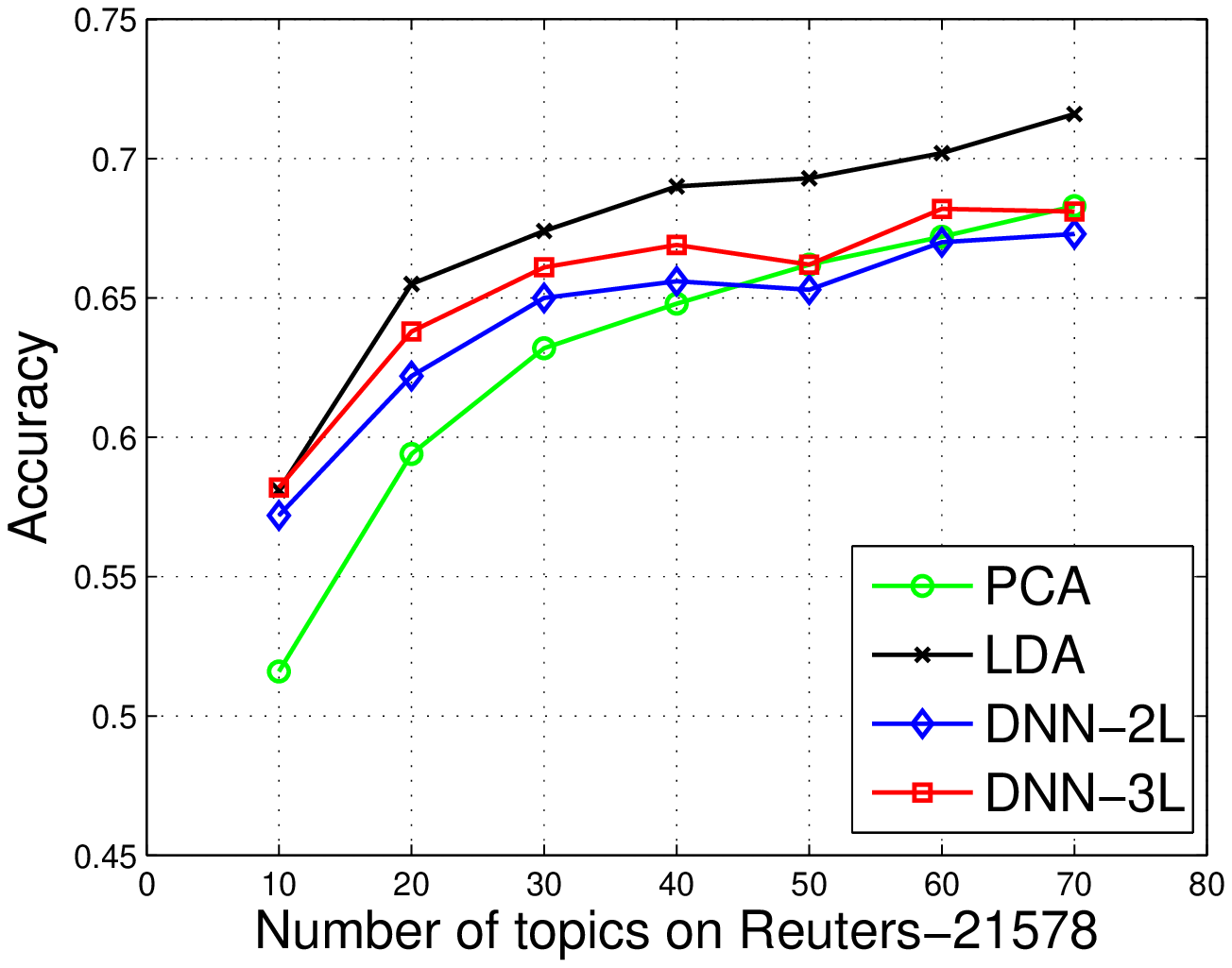} \\
\includegraphics[width=1\textwidth]{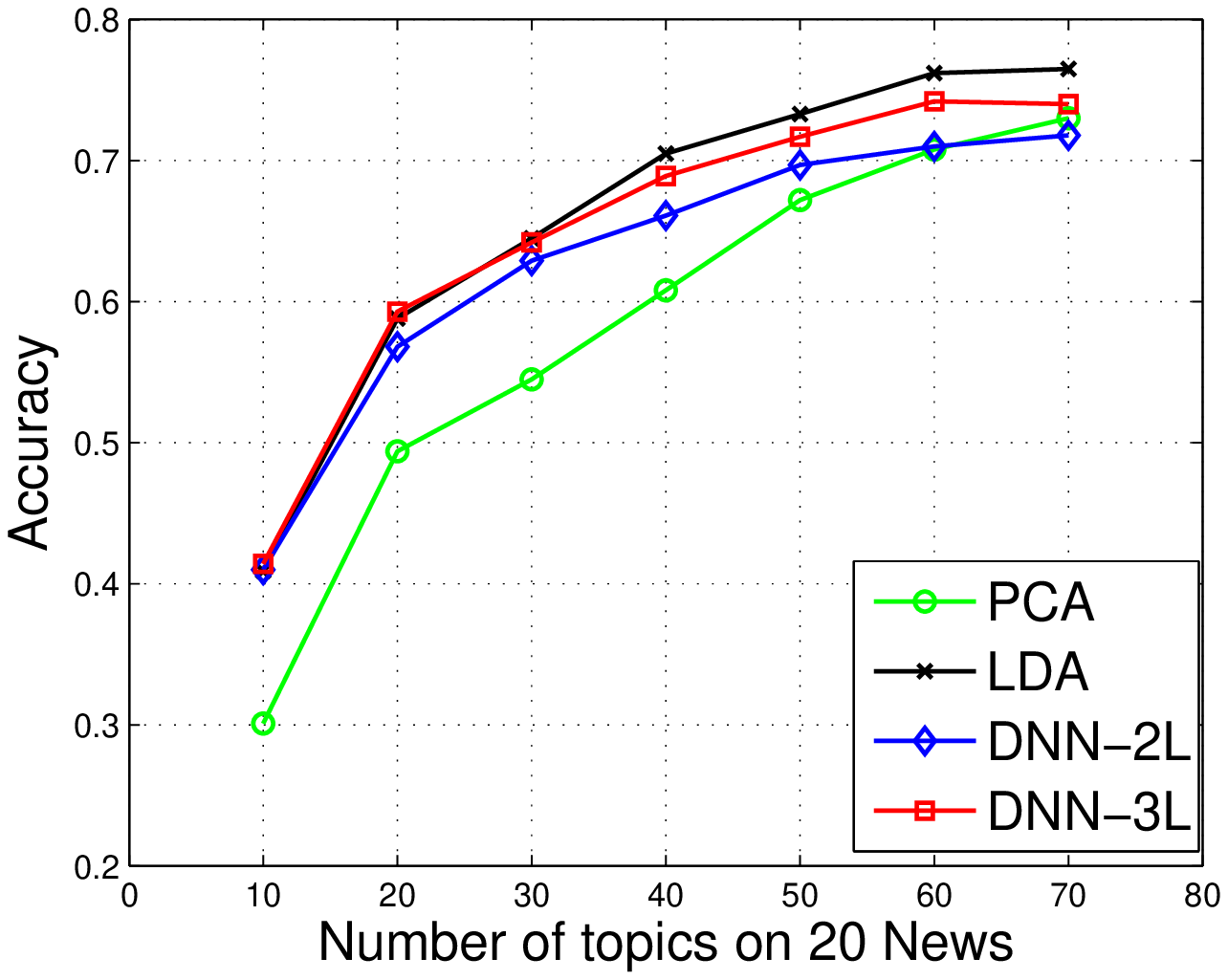} \\
\end{minipage}
}
\caption{The classification accuracy of PCA, LDA, 2-layer DNN (DNN-2L) and 3-layer DNN (DNN-3L).}
\label{fig:classification_accuracy}
\end{figure}
The results in terms of classification accuracy on the two datastes are reported in Figure~\ref{fig:classification_accuracy},
where the number of topics varies from $10$ to $70$. We first observe that LDA obtains better performance than PCA on
both the two datasets. Again, this is partly attributed to the long documents used in the study.
The two DNN models obtain similar performance as LDA and outperform PCA,
particulary with a small number of topics. This indicates that the DNNs indeed learned the behavior of LDA.
If the number of topics is large, the DNN models work not as well, possibly because the limited amount of training data (just several thousands of training samples) can not afford learning complex models.

\begin{figure}[!htb]
\centering
\subfigure{
\begin{minipage}[b]{0.4\textwidth}
\includegraphics[width=1\textwidth]{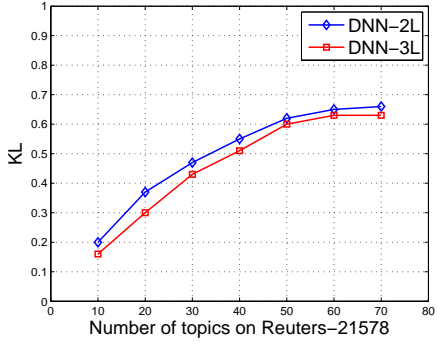} \\
\includegraphics[width=1\textwidth]{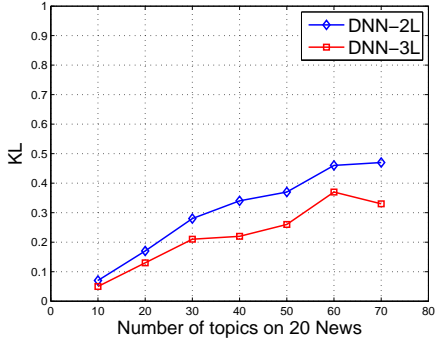} \\
\end{minipage}
}
\caption{The averaged KL divergency between the DNN and LDA output calculated on the test data of Reuters-21578 and 20 Newsgroups.}
\label{fig:KL}
\end{figure}

Note that the 3-layer DNN outperforms the 2-layer DNN. This indicates that deeper models
can learn the LDA behavior more precisely. This can be evaluated more directly in terms of
KL divergence between the LDA output $\theta(d)$ and the DNN prediction $f(v(d);w)$, as shown in Figure~\ref{fig:KL}.

%\begin{figure}[!htb]
%\centering
%\epsfig{figure=KL.eps,width=6cm,height=4cm}
%\caption{The averaged KL divergency between the DNN and LDA output calculated on the test data of Reuters-21578 and 20 Newsgroups. }
%\label{fig:kl}
%\end{figure}

\subsubsection{Inference speed}

%\begin{table}[!htb]
%\begin{center}
%\begin{tabular}{|l|c|c|c|}
%\hline
%\multirow{2}{*}{Model} & \multicolumn{2}{c|}{Seconds}        \\
%\cline{2-3}             &   Reuters  & 20 News                     \\
%\hline
%PCA                     &   75   &   76   \\
%LDA                     &   75   &   76   \\
%DNN (2 layers)          &   73   &  74    \\
%DNN (3 layers)          &   74   &  76   \\
%\hline
%\end{tabular}
%\end{center}
%\caption{\label{tab:inferTime} The inference time (s) for document vectors on the Reuters-21578 dataset.}
%\end{table}

\begin{figure}[!htb]
\centering
\epsfig{figure=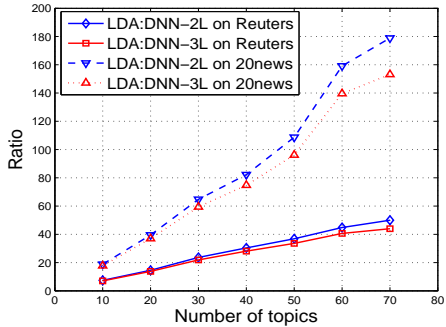,width=6.5cm,height=4.6cm}
\caption{The ratio of inference time of LDA to DNN. }
\label{fig:speed}
\end{figure}

The comparative results on inference time are shown in Figure~\ref{fig:speed}. The experiments were conducted on a desktop with 4 3.4G Hz cores, and to alleviate randomness the experiments were conducted $10$ times and the averaged numbers are reported. It can be seen that the DNN model is much faster ($10$ to $200$ times) than the original LDA, and the superiority is more clear with a large number of topics. Comparing the results
on the two datasets, we observe that DNN exhibits more advantages on 20 Newsgroups, because the long documents of this dataset are more difficult to infer with LDA. Additionally, the 3-layer DNN is not much slower than the 2-layer DNN, which means that using deeper models does not cause much additional computation.

\section{Topic discovery by transfer learning}
\label{sec:analysis}
%\begin{table}[!htb]
%\begin{center}
%\begin{tabular}{|l|c|c|}
%\hline
%\multirow{1}{*}{Dataset1} & Neutron    &  top tf-idf words        \\
%\cline{2-3}             &   Neutron    &  top tfidf words                    \\
%\hline
%Topic words   &        & seamen ship vessel     \\
%                      &        &  port strike Gulf    \\
%\hline
%\multirow{3}{*}{Dataset1} & Neutron    &  top tfidf words        \\
%   &   Neutron 3     &  loss profit \textbf{ship}    \\
%                      &        &  prices Inc Corp   \\
%Hidden layer 1   &   Neutron 43     &  quarter dlrs share    \\
%&        &  company \textbf{ship}  sales   \\
%   &   Neutron 45     &  quot corn offer    \\
%&        &  \textbf{ship}  money Bank   \\
%\hline
%Hidden layer 2   &  Neutron 45  &  \bf{seamen} port Gulf   \\
%   &    &  \bf{ship} strike vessel  \\

%\hline
%\end{tabular}
%\end{center}
%\caption{\label{tab:example1} The example of topic clustering of DNN.}
%\end{table}

\begin{figure}[!htb]
\centering
\epsfig{figure=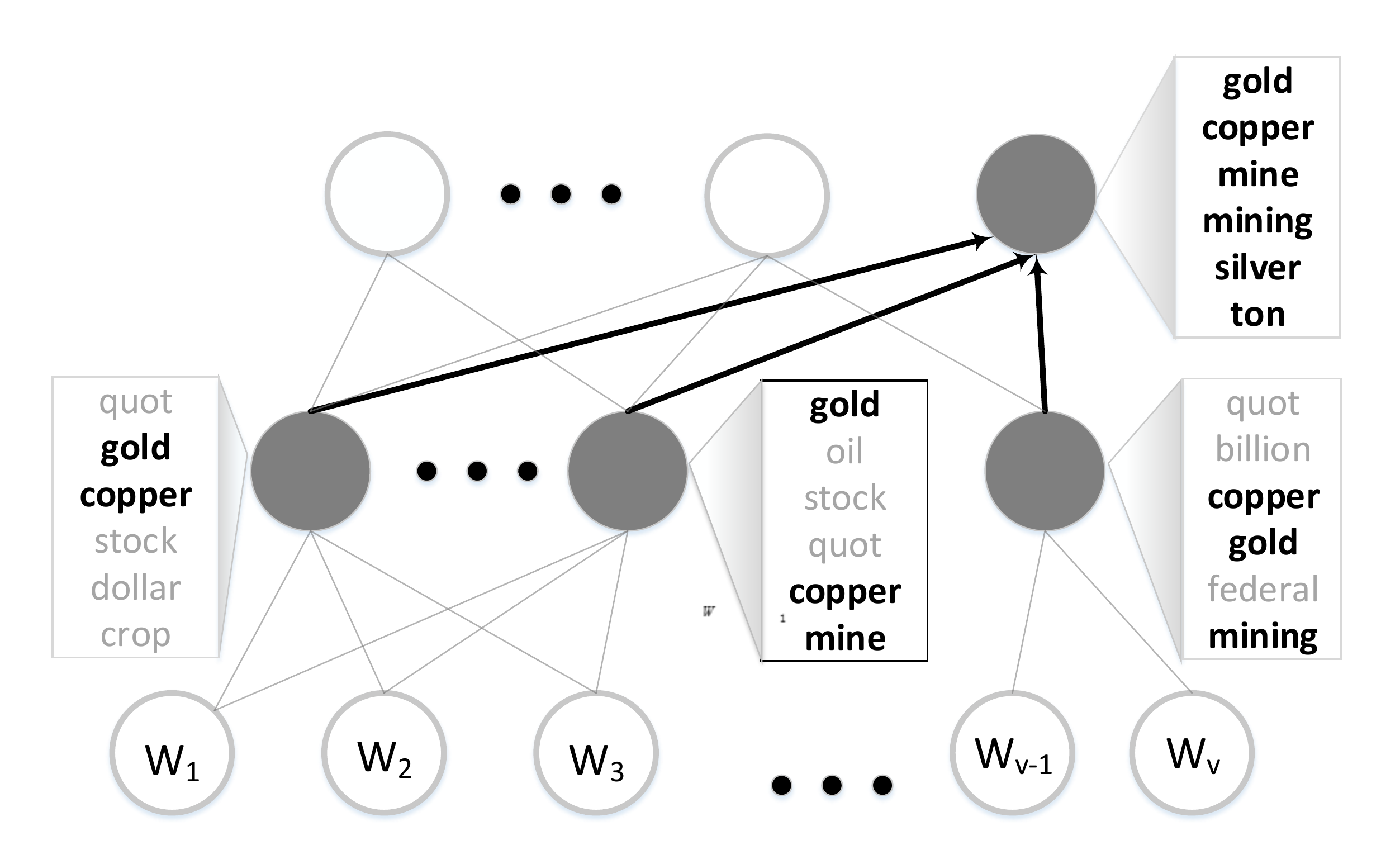,width=6cm}
\caption{Discovery for the topic `mining' with DNN. The words in dark are topic related words. }
\label{fig:exampleOfClustering_DNN}
\end{figure}

A known advantage of DNNs is that high-level representations can be learned automatically layer by layer. This
property may help DNN to discover topics from the raw TF input. To verify this conjecture, a one-hot vector is given to the DNN input,
and the activation on each hidden neuron is recorded. The one-hot vector represents a particular word, and
the activation reflects how a particular neuron is related to this word. For each neuron,
we record the activations of all the words and select the top-$10$ words that give the most significant activations, which
forms the set of representative words for the neuron.

Interestingly, we find that for each neuron, the representative words are generally correlated, forming a local topic.
%Topics of higher-level neurons are formed by topics of lower-level neurons, leading to a hierarchical
%topic clustering and discovery.
Figure~\ref{fig:exampleOfClustering_DNN} shows an example, where the topic `mining'
at the second hidden layer is formed by aggregating the related topics at the first hidden layer.
This example shows clearly how words are clustered layer by layer to form semantic meaningful topics.
Interestingly, we find that the topics derived from DNN and LDA are quite similar, and the DNN-derived topics look more
reasonable. As an example, the top-10 words for the topic `mining' derived from LDA are {\bf\{gold, said, mine, copper, ounces,
mining, tons, ton, silver, reuter\}}, while the DNN-derived top-10 words are {\bf \{gold, copper, mine, mining, silver, zinc,
minerals, metal, mines, ton\}}.

%This can be explained by the fact that the mixture weights generated by LDA and used as the DNN supervision are based on the
%same latent topics. We emphasize that the topic information is not transferred to DNN in the model training; it is the learning power of
%DNN that discovers the topics by itself.

%The dark colored words are topic related words. Its topic related words include "gold mine copper ounces mining tons ton silver production resources". After checking the news about this topic, we find the related news e.g. "Starting of gold production and mining tones of gold at Ketza river". %and "Hapag orders new ship from China and China will build the vessel".
%And we can not get the real topic from Hidden layer 1. The top tf-idf words in Neutron 6, Neutron 37 and Neutron 44 from Hidden layer 1 are not related to real topic. And we can get real related top tf-idf words in Neutron 38 from Hidden layer 2. And other topics also have similar properties as this topic. So we conclude that the hidden layer 2 represents the topic distribution of LDA. And each neutron represents a topic of LDA. And the layers and neutrons of DNN have explicable meanings.

\section{Conclusion and future work}
\label{sect:pdf}

We proposed a knowledge transfer learning method that uses deep neural networks to approximate LDA. Results on a document classification task show
that a simple DNN can approximate LDA quite well, while the inference is tens or hundreds of times faster.
This preliminary research indicates that transferring knowledge from Bayesian models to neural models is possible.
%This is quite promising and offers a new way to leverage the respective advantages of the two very different models.

The future work involves studying knowledge transfer between more complex probabilistic models and other neural models. Particularly,
we are interested in how to use the knowledge of probabilistic models to regularize neural models so that the neurons are more interpretable.

\newpage
% include your own bib file like this:
\bibliographystyle{acl}
\bibliography{reference}

\end{document}